\documentclass[mnsc,nonblindrev]{informs3} 

\OneAndAHalfSpacedXI

\usepackage{pgfplots}
\usepackage{graphicx}
\usepackage{float}
\usepackage{makecell}

\usepackage{algorithm2e} 

\usepackage{amsmath}
\usepackage{amssymb}
 
\usepackage{subcaption}

\usepackage{etoolbox}
\newtheorem{Theorem}{Theorem}
\newtheorem{proposition}{Proposition}
\usepackage{natbib}
\usepackage{hyperref} 
\hypersetup{colorlinks=true,allcolors=[rgb]{0.5, 0.0 , 0.13}}
\begin{document}



\RUNTITLE{Enhancing Hospital Capacity Management}

\TITLE{ \Large Enhancing Uncertain Demand Prediction in Hospitals  \\ Using Simple and Advanced Machine Learning}

\ARTICLEAUTHORS{
\AUTHOR{\small Annie Hu*, Samuel Stockman*} 
\AFF{School of Mathematics, University of Bristol, UK.} 

\AUTHOR{\small Xun Wu}
\AFF{School of Engineering and Informatics, University of Sussex, UK} 

\AUTHOR{\small Richard Wood}
\AFF{NHS Integrated Care Board, UK} 

\AUTHOR{\small Bangdong Zhi$\dagger$}
\AFF{School of Business, University of Bristol, UK.} 

\AUTHOR{\small Oliver Y. Ch\'en$\dagger$}

\AFF{Centre Hospitalier Universitaire Vaudois (CHUV) and Université de Lausanne, Switzerland} 

} 

\ABSTRACT{%
Early and timely prediction of patient care demand not only affects effective resource allocation but also influences clinical decision-making as well as patient experience. Accurately predicting patient care demand, however, is a ubiquitous challenge for hospitals across the world due, in part, to the demand's time-varying temporal variability, and, in part, to the difficulty in modelling trends in advance. To address this issue, here, we develop two methods, a relatively simple time-vary linear model, and a more advanced neural network model. The former forecasts patient arrivals hourly over a week based on factors such as day of the week and previous 7-day arrival patterns. The latter leverages a long short-term memory (LSTM) model, capturing non-linear relationships between past data and a three-day forecasting window. We evaluate the predictive capabilities of the two proposed approaches compared to two naïve approaches - a reduced-rank vector autoregressive (VAR) model and the TBATS model. Using patient care demand data from Rambam Medical Center in Israel, our results show that both proposed models effectively capture hourly variations of patient demand. Additionally, the linear model is more explainable thanks to its simple architecture, whereas, by accurately modelling weekly seasonal trends, the LSTM model delivers lower prediction errors. Taken together, our explorations suggest the utility of machine learning in predicting time-varying patient care demand; additionally, it is possible to predict patient care demand with good accuracy (around 4 patients) three days or a week in advance using machine learning. 


}%


\KEYWORDS{predictive capability; patient care demand; time-varying linear approach; long short-term memory approach} \HISTORY{Current version: April 2024.}

\maketitle

\section{Introduction}\label{sec: introduction}
The healthcare industry is a vital component of human society, ensuring the provision of essential healthcare services to communities. Within this sector, hospitals serve as primary institutions responsible for delivering these critical services. Hospitals have experienced uncertain demand across multiple aspects of patient care, attributable to factors such as population growth, changing demographics, and an increasing prevalence of individuals living with multiple comorbidities \citep{de2023dynamic}. 

As a result of these uncertainties in demand, hospitals face significant challenges in managing their limited available capacity, which in turn compromises the quality of care provided to patients. To address these pressures, hospitals have begun integrating forecasting approaches into their operations \citep{wang2024wasserstein}. This study aims to investigate the potential of new approaches to better manage the multifaceted surge in patient demand within hospitals, taking into account the distinct time frames and nature of patient care demand.

Using data from the Rambam Medical Center, an academic hospital in Israel that serves 75,000 patients annually from a surrounding population of over two million. We have demonstrated the forecasting capability of two proposed models.\footnote{For forecasts to be useful for resource management within hospitals, a minimum 3-day horizon is required to give sufficient preparation time.} The first, a time-varying linear model, makes linear predictions on the number of patients arriving at each hour of a week based on a set of features that include the day of the week and the last 7-day hourly number of arrivals. The second model, a neural network known as an LSTM, learns a non-linear relationship between the previous week of data and a three-day forecasting horizon. Both models demonstrate an ability to capture hourly variability, however, the LSTM model better describes weekly seasonal effects and therefore achieves the lowest prediction error of the two. 

This paper contributes to forecasting the demand of healthcare industry mainly in two ways. First, many previous studies focus on Bayesian methods to solve prediction problems. For example, \citet{brahim2004} propose a Gaussian process model for non-stationary time series with hyper-parameters being estimated by maximising the log-likelihood function using a training dataset. This Gaussian process forecasting method has been shown to outperform radial basis function (RBF) neural network. \citet{sosa2022} propose a forecasting approach based on linear time-varying regression. Radial basis function is applied to approximate the dynamic parameters in the non-parametric linear regression model. However, Bayesian based approaches have some limitations. For instance, Bayesian methods can be computationally expensive when the data size $T$ is very big due to an inversion of a large matrix, which requires the memory to be $\mathcal{O}(T^2)$ to store the matrix, and the computational complexity is $\mathcal{O}(T^3)$ for inverting the matrix, and some approaches require batch learning, which might not be able to capture changes as time varying. To overcome these difficulties, our proposed time-varying model has a state-space model form, and Kalman filter can be applied to recursively make predictions on this model, which can be more efficient and adaptive better with time changing. It can also be shown that Kalman filter us a recursive Bayesian filter.
Second, our paper introduces a machine learning approach utilizing the Long Short-Term Memory (LSTM) recurrent neural network architecture. The LSTM model learns the non-linear dependencies between $k$-day hourly forecasts and the previous week's arrival counts. Through an iterative updating procedure, the model learns a vector representation of the historical arrival counts. This vector is then used to make $k$-day forecasts via a one-layer fully connected neural network. The LSTM approach also allows for the incorporation of external features, such as weather data, to improve forecast accuracy. By utilizing hourly maximum temperature as an additional feature, our LSTM model enhances its predictive capabilities. Our proposed LSTM approach offers a robust and efficient method for forecasting healthcare demand, addressing key limitations of previous Bayesian methods and providing valuable insights for healthcare management and resource allocation.

This paper is structured as follows: Section \ref{sec: related literature} reviews related literature aimed at enhancing predictive performance in forecasting patient care demand. In Section \ref{sec: EDA}, a descriptive analysis of data from the Rambam Medical Center is presented. Section \ref{sec: forecasting model} outlines the predictive models employed in this study. Finally, Section \ref{sec: conclusion} provides a comprehensive conclusion, along with insights into potential future research avenues.

\section{Related Literature} \label{sec: related literature}
This study is closely related to the literature that intends to improve the predicative performance for patient arrivals. Most notably, \cite{whitt2019forecasting} benchmark forecasting models for patient arrivals on the same dataset as this study. In their experiment, a seasonal autoregressive integrated moving average with exogenous (holiday and temperature) regressors (SARIMAX) time-series model is found to achieve the best forecasting performance out of several other models that include a neural network approach. A comparison of the models is made by forecasting daily arrival counts at a one-day horizon. Following that, an hourly forecasting model is constructed from the SARIMAX model by constructing a non-homogeneous Poisson process. However, the forecasting horizon does not extend beyond 6 hours. Furthermore, the machine learning approach that they adopt (the multi-layer perceptron) is not designed for sequential data. Instead, networks with sequential structures such as Recurrent Neural Networks (RNN) are more competitive in such instances \cite{hewamalage2021recurrent}. \cite{sudarshan2021performance} and \cite{notz2023prescriptive} also construct a forecasting experiment that compares models that forecast hospital arrivals. Specifically, the suite of models in \cite{sudarshan2021performance}  includes an LSTM as in this study, however, the forecasts are for daily arrivals and do not describe any variation across the course of a single day. Compared with the aforementioned literature, considering the aforementioned factors, we explore the productivity performance of new approaches. Compared with \cite{hewamalage2021recurrent},  a 6-hour forecast model, our models have better predictive performance, much lower than their reported MSE of 160.16. 

\section{Exploratory Data Analysis}\label{sec: EDA}
We use data from the Rambam Medical Center, an academic hospital in Israel that serves 75,000 patients annually from a surrounding population of over two million. The dataset tracks the paths of individuals throughout their stay at the hospital including admission, discharge and transfers between two of the 45 medical units. The dataset covers the period 01/01/2004 -  01/11/2007 and is known to be complete. The time period, however, contains the 2006 Lebanon war, the war between Israel and Lebanon from
 12/07/06 to 14/08/06. During the war, patient arrivals vary significantly from the remainder of the data. For that reason we ignore the year 2006 in our forecasting experiment completely.

We begin with a brief exploratory analysis of the arrival times of patients to the hospital. Even though the dataset tracks the arrivals of patients to different departments (60\% to the Emergency Department 20\% directly to the Wards, and 20\% to the X-Ray department), we treat all departments as one whole. Since we use the year 2007 for testing our models, data from this year is excluded from the analysis.

\begin{figure}[h!]
\centering
\begin{subfigure}[b]{0.8\textwidth}
    \centering
   \includegraphics[width=0.8\textwidth]{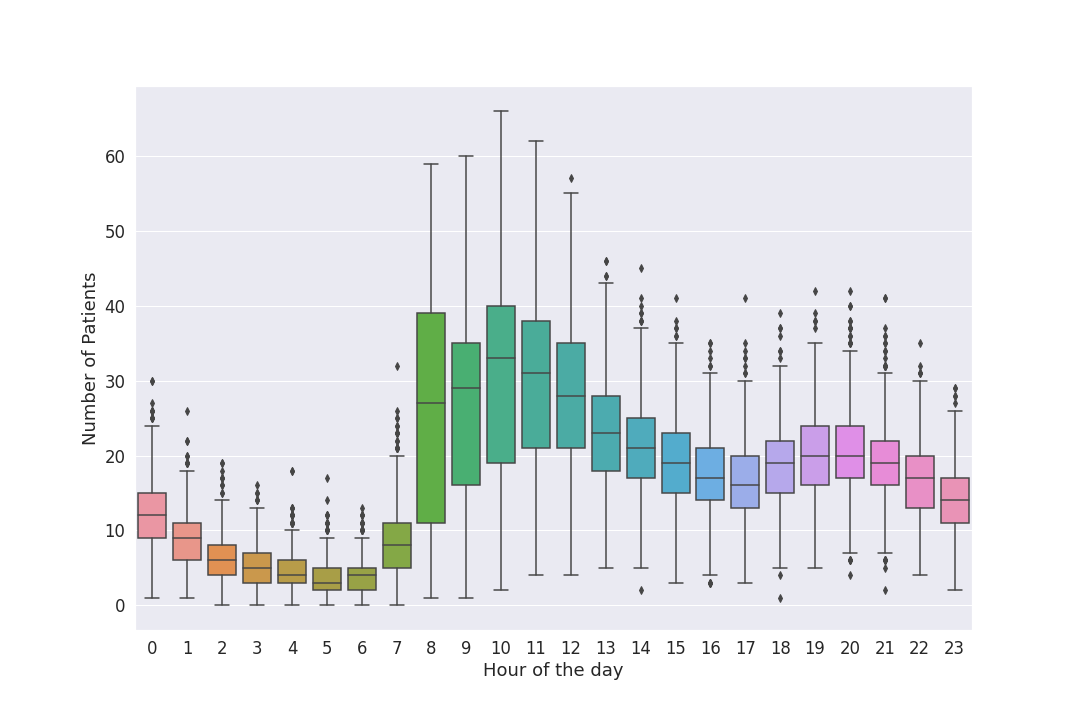}
   \caption{}
   \label{fig: hourly} 
\end{subfigure}
\begin{subfigure}[b]{0.8\textwidth}
    \centering
   \includegraphics[width=0.8\textwidth]{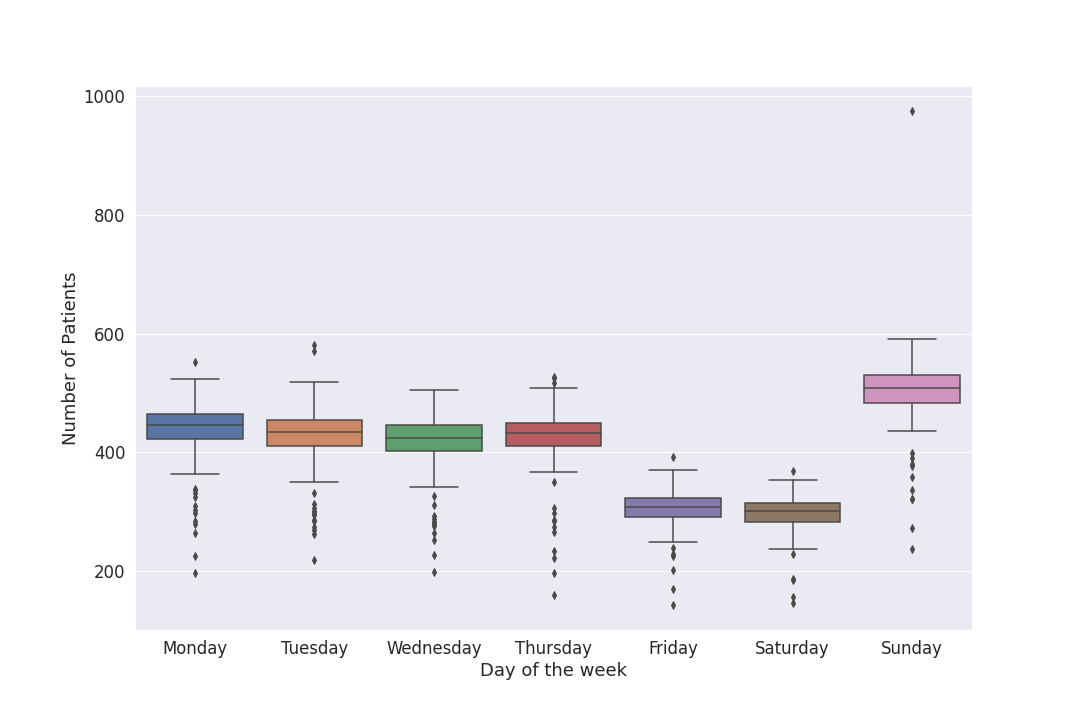}   
   \caption{}
   \label{fig:DOW}
\end{subfigure}
\caption{(a) Boxplot of the number of patients arriving at each hour in the day to the Rambam Medical Center.(b) The number of patients arriving on each day of the week.}
\end{figure}

First, patient arrival times are aggregated so that we have counts of how many patients arrive hourly. Figure \ref{fig: hourly} shows a boxplot of the number of patients arriving at each hour of the day. Understandably, the first seven hours of the day see the lowest arrival counts until 08:00 where a surge of arrivals occur. There is a second small surge of arrivals around 19:00-20:00.

\begin{figure}
    \centering
    \includegraphics[width=\textwidth]{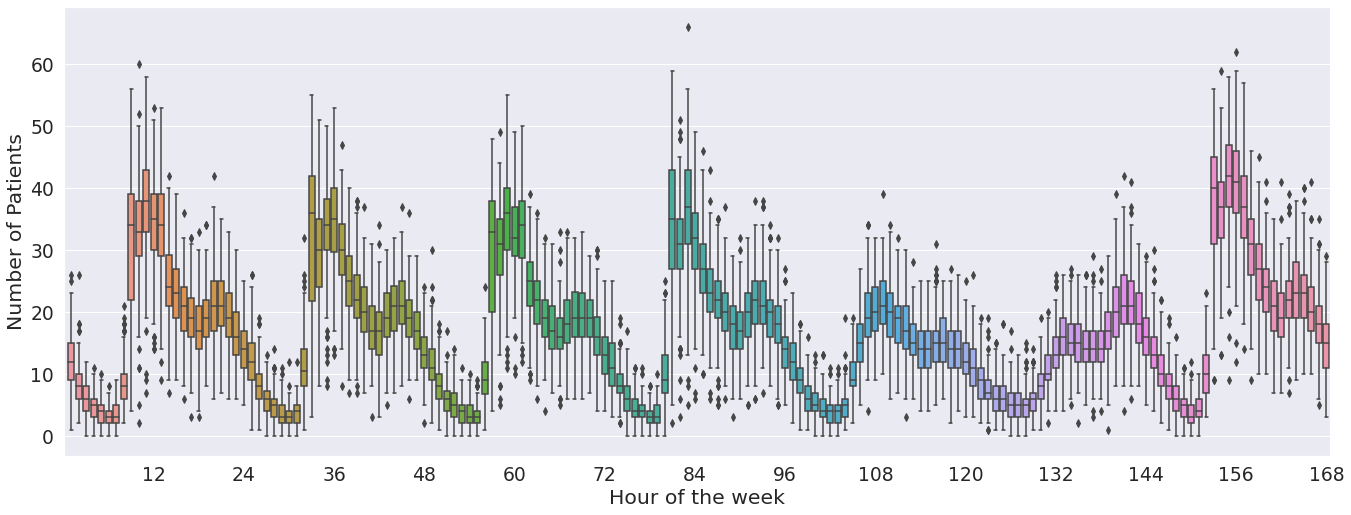}
    \caption{Boxplot of the number of patients arriving at the Rambam Medical Center at each hour of the week beginning 00:00 Monday.}
    \label{fig:HOW}
\end{figure}

Figure \ref{fig:DOW} shows the number of patients arriving at each day of the week. Since the Rambam Medical Center is in Israel, the drop in the number of patients arriving on Friday and Saturday reflects the Israeli weekend (Friday-Saturday). The first day of the working week (Sunday) sees a slightly higher than average patient count than other weekdays. The weekend days do not however, have the same hourly shape as the weekdays, Figure \ref{fig:HOW}. The greatest period of inactivity is over the religious holiday, Shabbat, which lasts from Friday sundown to Saturday sundown. There are small surges immediately before and after the holiday.

\begin{figure}
\centering
\begin{subfigure}[b]{0.9\textwidth}
    \centering
   \includegraphics[width=0.9\textwidth]{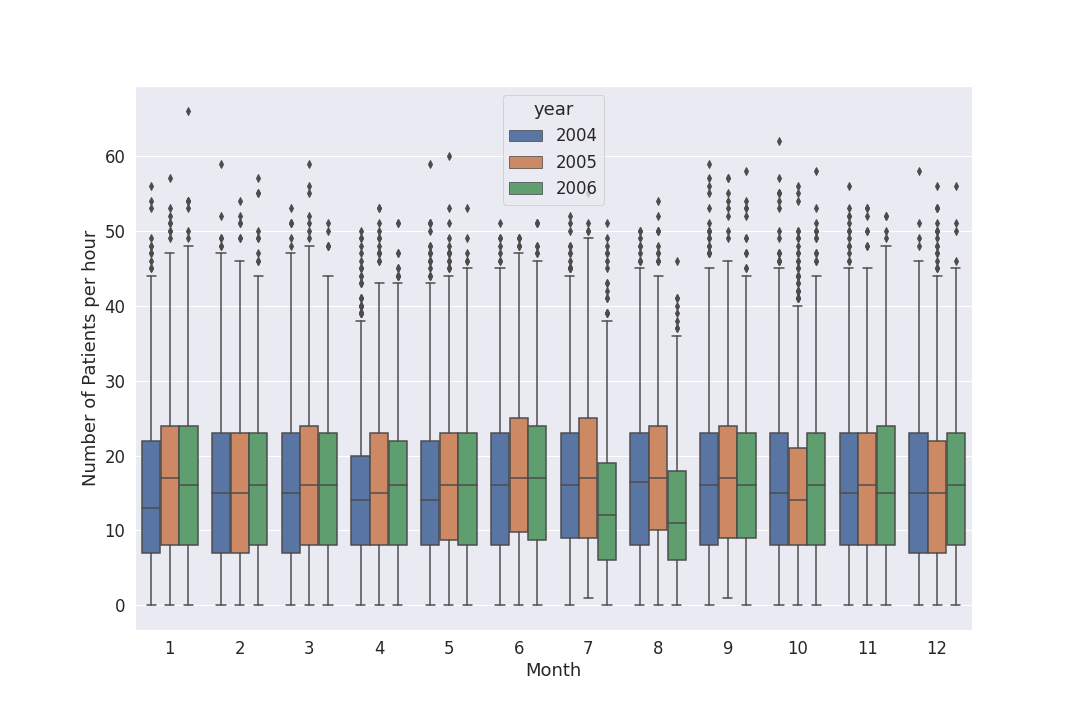}
   \caption{}
   \label{fig:monthly} 
\end{subfigure}
\begin{subfigure}[b]{0.9\textwidth}
    \centering
   \includegraphics[width=0.9\textwidth]{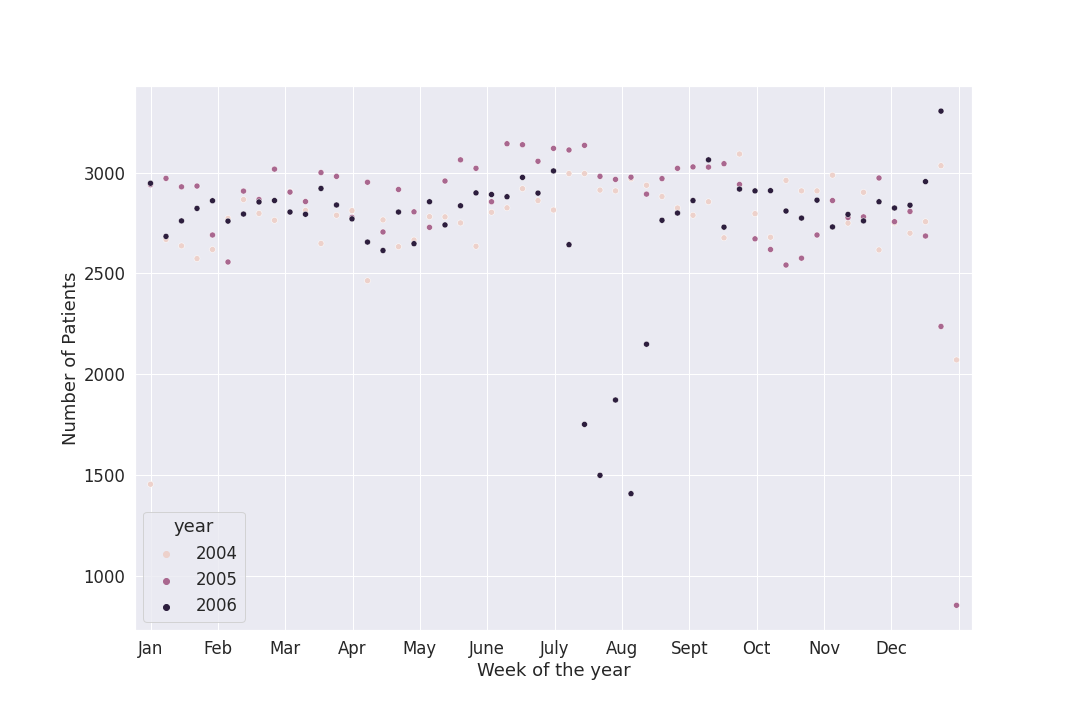}
   \caption{}
   \label{fig:WOY}
\end{subfigure}
\caption{(a) Boxplot of the number of patients arriving at the Rambam Medical Center per hour for each month of the year. (b) The number of patients that arrive on each week of the year.}
\end{figure}

There is very little variation in the patient arrival rate over the course of the year, Figures \ref{fig:monthly} and \ref{fig:WOY}, with the only deviation coming from the two month period in 2006 during the Lebanon war.

\section{Forecasting Model} \label{sec: forecasting model}
We now use the hourly count data in an experiment to compare two forecasting models with the other three models: the naive models, which acts as a benchmark, the reduced-rank vector autoregression model, and the TBATS model. 
\subsection{Naive Approach}
Since we have observed that the number of arrivals of a year is rather periodic on a week basis, we use a naive method \citep{paldino2021does} as a baseline model to measure how well our models can perform. The naive method we are using is to assume the forecasts of arrivals for the week $t+1$ are just the actual numbers of arrivals for the week $t$. The MAE of this naive method is about $4.94$, which illustrates that our models can be around $1$-person more accurate per hour than the naive method. 
\subsection{Reduced-rank Vector Autoregression Model (RVAR)}
The RVAR models \citep{velu1986reduced} have the following form:
\begin{equation}
    y_t  = c+A_1y_{t-1}+A_2y_{t-2}+\dots+A_py_{t-p}+e_t,
\end{equation} where $y_1, y_2, \dots, y_T\in\mathbb{R}^{n\times 1}$ are observations at each time $t\in\{1, 2, \dots, T\}$, $p$ is the number of previous observations up to time $t-1$ we use to predict $y_{t}$, and $e_1, e_2, \dots$ are the error terms, which are independent of each other and with an expectation $0$. The RVAR method is based on the VAR method. For example, suppose $X_1,\ldots, X_T$ is a times series of length $T$, and  $X_t\in\mathbb{R}^N$ for all $t\in\{1,\ldots, T\}$, then we can  write the VAR(1) model as following:
\begin{equation}
    X_t  = AX_{t-1}+e_t.
\end{equation} The matrix $A\in\mathbb{R}^{N\times N}$ can be expressed as $A=WV$, where $W\in\mathbb{R}^{N\times R}$ and $V\in\mathbb{R}^{R\times N}, 1\leq R\leq N$. By imposing a suitable reduced rank $R$, the number of parameters in matrices $W$ and $V$ can be less than the number of parameters in matrix $A$, which can help with the over-parametrisation problem of the VAR method. Let $X_2=[y_2, y_3, \dots, y_T]$ and $X_{1}=[y_1, y_2, \dots, y_{T-1}]$, then the matrices $W$ and $V$ can be solved by minimising the following function:
\begin{equation}
    \operatorname{min}_{W,V} \frac{1}{2}\|X_2-WVX_1\|_{\operatorname{F}}^2,
\end{equation} where $\|\cdot\|_{\operatorname{F}}$ denotes the Frobenius norm. Then we have $$W=X_2(VX_1)^{-1}, \quad V = W^{-1} X_2X_1^{-1},$$ provided that we have initialised $V$ with some random value.  We use the number of arrivals of week $t$ and $t-1$ to predict patient arrivals of the week $t+1$, which shows an MAE of $4.96$, higher than the time-varying linear model and the LSTM model.
\subsection{Time-varying Linear Model}
 We propose an approach based on state space model and Kalman filter, which is a recursive Bayesian filter for multivariate normal distributions. The proposed approach is a time-varying linear Gaussian model that has a state-space model representation \citep{hamilton1994state}. Kalman filter is then applied to make forecasts, and update the parameters in the state space model with time changing \citep{arnold1998, koopman1997}. The proposed method is an online learning algorithm. That is, the model learns and updates the parameters every time we receive a new observation, and then a prediction is made for the next observation, which requires less computational cost than standard Bayesian method.

\citet{guo1990estimating, zhang1991lp} propose a Kalman filter based estimator for stochastic linear regression models with time-varying parameters:
\begin{equation}
    \begin{cases}
        &y_i = \beta^\top(t_i) x_i+\epsilon_i \quad \epsilon_i\sim\mathcal{N}(0,R)\\
        \vspace{-0.4cm}
        \\
        &\beta_i = \beta(t_{i-1})+v_i, \quad v_i\sim\mathcal{N}(0,Q)
    \end{cases}\quad i \geq 1,
\end{equation} where \citet{guo1990estimating} illustrates the convergence and stability properties of the estimator $\hat{\beta}$. 
The paper guarantees that, when our interested parameters are time-varying, and the tracking error is bounded, then using the Kalman filter algorithm, we have that both:\\
(1) $\limsup\limits_{n\rightarrow\infty}\mathbb{E}[\|\hat{\beta}_{n}-\beta_{n}\|^2]$ and,\\
(2) $\limsup\limits_{n\rightarrow\infty}\frac{1}{n}\sum_{i=0}^{n-1}\|\hat{\beta}_{i}-\beta_{i}\|$\\
are bounded a.s.

In our model, we assume that there is a hidden process $\beta({t_i})\in\mathbb{R}^d$ and observations $y_i,\, i = 1, 2, \dots, T$, and the model is of the following state-space model form:

\begin{equation}\label{eq:ssm}
\begin{cases}
    &\beta(t_0)\sim\mathcal{N}(\mu_0, V_0), \\
    \vspace{-0.4cm}
        \\
    &\beta(t_i)=B \beta(t_{i-1})+\epsilon_{1,i-1},\\
    \vspace{-0.4cm}
        \\
    &Y_{i}=H_{i}\beta(t_i)+\epsilon_{2,i},
    \end{cases}
\end{equation} with the initial $\beta(t_0)\sim\mathcal{N}(\mu_0, V_0)$, where $\mathbb{E}[\epsilon_{2,i}]=0$, $\mathbb{E}[\epsilon_{2,i}^2]=\sigma^2 $, $\mathbb{E}[\epsilon_{1,i}]=0$ and $\mathbb{E}[\epsilon_{1,i}\epsilon_{1,i}^\top]=Q=\operatorname{diag}(\sigma^2)$, $H_i$ represents the predictors, which is a row vector containing the factors that have influences on the observations, and $B = \operatorname{diag}(\alpha_1,\dots,\alpha_d).$ 
\begin{Theorem}
\label{marginal}
Let $X$ and $M$ be random variables with distributions: $p(X|M = m)=\frac{1}{\sigma\sqrt{2\pi}}e^{-\frac{1}{2}\frac{(X-m)^2}{\sigma^2}}$, and $p(M)=\frac{1}{s\sqrt{2\pi}}e^{-\frac{1}{2}\frac{(M-\theta)^2}{s^2}}$, then the marginal distribution $p(X)=\mathcal{N}(X;\theta,s^2+\sigma^2)$.
\end{Theorem} 

Let $\mathbb{E}[\beta(t_i)\,|y_1, \ldots,y_i]=\mu_{i\,|i},\,\operatorname{Cov}[\beta(t_i)\,|y_1, \ldots,y_i]=V_{i\,|i}$. We can then obtain the marginal likelihood function $p(y_{1},...,y_{T})$ by using Kalman filter update:
\begin{equation}
    \begin{aligned}
      &\mu_{i|i} = B\mu_{i-1|i-1} + G_{i} [y_{i} - H_iB\mu_{i-1|i-1} ],\\
      &V_{i|i} = P_{i|i-1} - G_{i} H_i P_{i|i-1},
    \end{aligned}
\end{equation} where
\begin{equation}
    P_{i|i-1}=BV_{i-1|i-1}B^\top + Q
\end{equation} and
\begin{equation}
    G_{i}=P_{i|i-1}H_i^\top(H_iP_{i|i-1}H_i^\top+\sigma^2)^{-1}, i=1,2,...,T.
\end{equation}
\begin{proposition}
\label{prop1}
By Theorem \ref{marginal}, we have 
$$
\begin{aligned}
    p(y_{1},...,y_{n})&=\mathcal{G}(y_{1};H_1B\mu_0,\sigma^2+ H_1(BV_0B^\top+Q)H_1^\top)\\
    &\prod_{i=2}^n\mathcal{G}(y_{i}; H_iB\mu_{i-1|i-1}, \sigma^2+
    H_iP_{i|i-1}H_i^\top),
\end{aligned}
$$ where $\mu_0$ and $V_0$ are the initial mean and covariance of $\beta(t_0)$ respectively, and $P_{i|i-1}$ is the updated covariance of $\beta(t_i)$ given $y_{1}, \dots y_{i-1}$ by Kalman filter, and $\mathcal{G}$ denotes the Gaussian distribution.
\end{proposition}

We consider $y_{i}$ as a vector composed of 7-day (Monday to Sunday) hourly arrivals, and we predict for the next week's hourly arrivals, $\hat{y}_{i+1}$. As we can see, the model can adapt with new data appearing, and we can use Kalman filter to update the parameters $$\hat{\mu}_{i+1} = \mu_{i+1|i} = B\mu_{i|i},$$ and $$\hat{V}_{i+1} = P_{i+1|i}.$$ We convert the data to an hourly dataset which summarises the number of patients arriving at the hospital for every hour interval during 2004-2005 to make it a training dataset to estimate the hyper-parameters $(\alpha_1, \dots, \alpha_d, \sigma)$ by using grid search to optimise the marginal likelihood function. The likelihood function can be obtained by the Gaussian distribution provided in Proposition \ref{prop1}. We can then forecast the hourly arrivals of 7 days at once for the first ten months of 2007 by using week $t-1$'s data to predict week $t$'s arrivals. As shown in the previous data analysis section, we observe that the arrivals of patients at the hospital can be influenced by the day of the week, so we take account of the day of a week, Monday to Sunday as features in the linear model. 
\subsection{TBATS Model}
Since our data demonstrates multiple seasonal components (daily and weekly variation), we choose to benchmark a popular model for time-series with complex seasonality. TBATS (Trigonometric seasonality, Box-Cox transformation, Autoregressive Moving Average (ARMA) errors, Trend, and Seasonal components) models are state-space models that use a trigonometric  representation  of seasonal  components  based  on  Fourier  series. The TBATS state-space model can be summarised by the following equations,\newline Box-Cox transformation,
\begin{equation}
    y^{(\omega)}_t = \begin{cases}
    \frac{y^\omega_t-1}{\omega}& \text{if } \omega \neq 0;\\
    \log y_t,              & \text{if } \omega=0
\end{cases}
\end{equation}
Seasonal Periods,
\begin{equation}
    y_t^{(\omega)} = l_{t-1}+ \Phi b_{t-1} + \sum_{i=1}^M S^{(i)}_{t-m_i}+d_t
\end{equation}
Global and local trend,
\begin{align}
    &l_t = l_{t-1} + \Phi b_{t-1} + \alpha d_t \\
    &b_t = (1-\Phi)b+\Phi b_{t-1}+\beta d_t
\end{align}
ARMA error,
\begin{equation}
    d_t = \sum_{i=1}^p \Phi d_{t-i} + \sum_{j=1}^q \theta_j \epsilon_{t-j} + \epsilon_t
\end{equation}
Fourier seasonal terms,
\begin{align}
    &S_t^{(i)} = \sum_{j=1}^{k_i}S^{(i)}_{j,t}\\
    &S^{(i)}_{j,t} = S^{(i)}_{j,t-1}\cos \lambda_j^{(i)} + S^{*(i)}_{j,t-1}\sin \lambda_j^{(i)} + \gamma_1^{(i)}d_t \\
    &S^{*(i)}_{j,t} = -S^{(i)}_{j,t-1}\sin \lambda_j^{(i)} + S^{*(i)}_{j,t-1}\cos \lambda_j^{(i)} + \gamma_2^{(i)}d_t 
\end{align}
where $y_t$ denotes the hourly arrival count, $y_t^{(\omega)}$ represents the Box-Cox transformation of $y_t$ with parameter $\omega$, $l_t$ is the local level, $b_t$ the short term trend, $\Phi$ the trend damping parameter, $S_t^{(i)}$ the $i$th seasonal component, $m_1,\dots, m_T$ are the seasonal periods, $d_t$ represents an ARMA(p,q) process (\cite{brockwell2002introduction}), $\epsilon_t$ is a Gaussian white noice process, $\alpha, \beta$ and $\gamma_i$ are smoothing parameters, $\lambda_j^{(i)} = 2\pi j/m_i $ is the stochastic level of $i$th seasonal component, $S^{*(i)}_{j,t}$ is the change in the stochastic of $i$th seasonal component over time, and finally $k_i$ is the required number of harmonics for the $i$th seasonal component. For a detailed explanation of the model see \cite{de2011forecasting}.
\subsection{Long Short-Term Memory}
The final model is a machine learning approach using the Long Short-Term Memory (LSTM) recurrent neural network architecture. In this model, $k$-day hourly forecasts $\{y_{t+1},\dots, y_{t+k\times24}\}$ are non-linearly dependent upon the previous week of arrival counts $\{ y_{t},\dots,y_{t-168}\}$. The neural network learns a vector representation of the history of arrival counts $\textbf{h}_{t-i}$ through an iterative updating procedure,
\begin{equation}
    \textbf{h}_{t-i} = LSTM(\textbf{h}_{t-i-1},y_{t-i-1}) \ \ \text{for } i=167,\dots,0,
\end{equation}
where the vector $\textbf{h}_{t-i}$ is updated based on the observed count $y_{t-i-1}$ and last state of the vector $\textbf{h}_{t-i-1}$. For a detailed explanation on the form of the update see \cite{hochreiter1997long}. $k$-day forecasts are then made by learning a prediction function, 
\begin{equation}
    \{ y_{t+1},\dots,y_{t+k\times24}\} = f_\theta(\textbf{h}_{t}),
\end{equation}
that depends on this vector representation at the most recent time-step $t$. This prediction function is a one-layer fully connected neural network. The weights in the network are learnt through Stochastic Gradient Descent (SGD) of the loss function:
\begin{equation}
    \min_\theta \sum_{t} (y_t- f_\theta(\textbf{h}_t))^2.
\end{equation}
This approach has the capability of adding external features for use in the forecasts. In this case, the LSTM updating procedure is instead,
\begin{equation}
    \textbf{h}_{t-i} = LSTM(\textbf{h}_{t-i-1},\mathbf{x}_{t-i-1}) \ \ \text{for } i=167,\dots,0,
\end{equation}
where now $\mathbf{x}_{t-i-1}$ is the feature vector that also includes the count at each hour. We adopt the same external feature approach as \cite{whitt2019forecasting} by using weather data from the nearest weather station to the Rambam Medical Center. Their study found that temperature was the only informative feature to forecasts. In light of that we use the hourly maximum temperature as an additional feature in the LSTM model.

\subsection{Results}
The forecasts of patients arriving at the Rambam Medical Center were generated using various models, including time-varying linear models, TBATS, and different configurations of LSTM models.  Models are trained on data from  01/01/2004 - 31/12/2005 and the models are evaluated on data from 01/01/2007 - 01/11/2007. The linear method forecasts 7 days of hourly arrival counts into the future and the LSTM model 3 days of hourly arrival counts. We then compare the models using mean squared error and mean absolute error.

\begin{figure}
    \centering
    \includegraphics[width=0.8\textwidth]{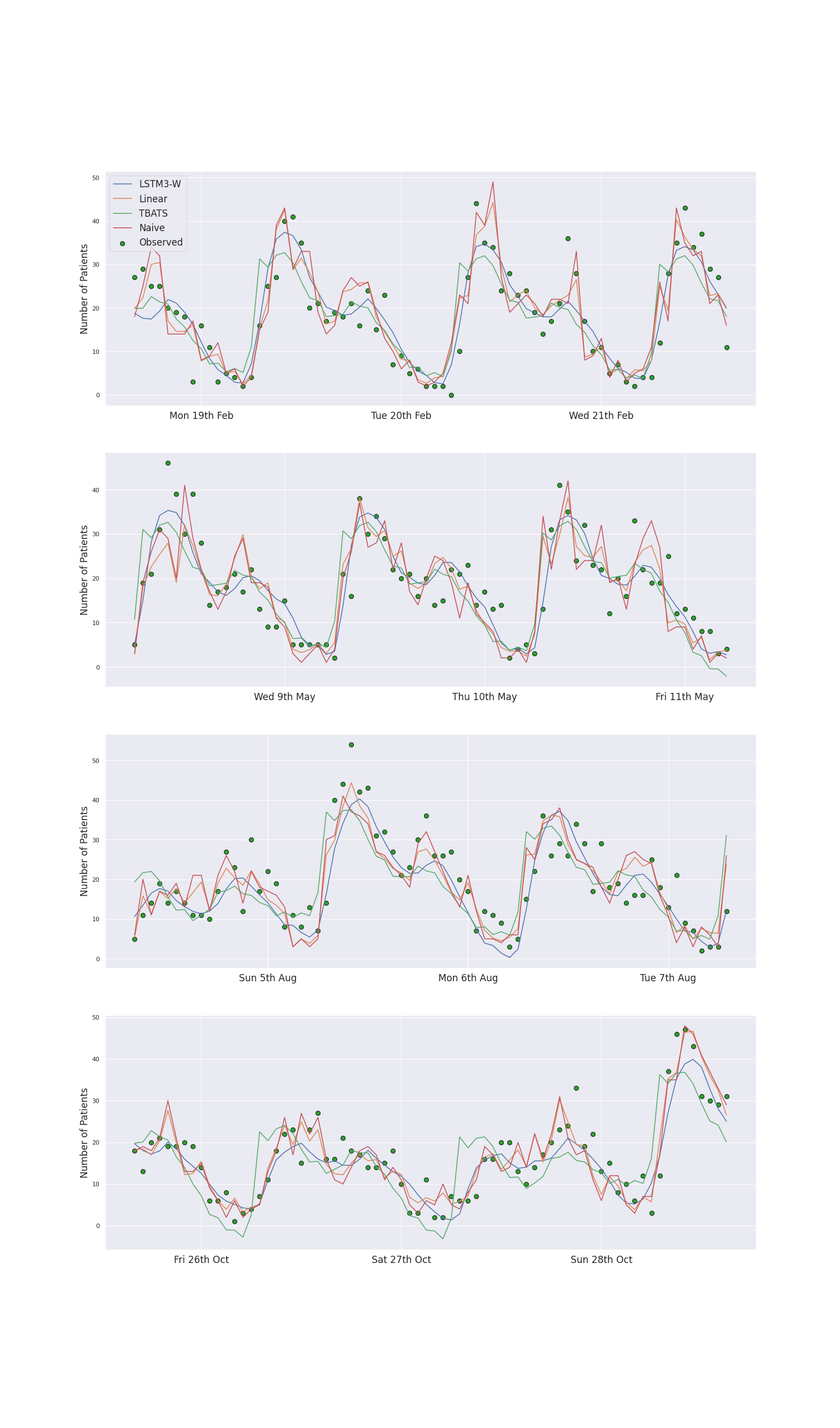}
    \caption{Forecasts of patients arriving at the Rambam Medical Center by both the LSTM and linear model. Blue forecasts from the LSTM model and orange from the linear model are compared with the true observed points in green.}
    \label{fig:forecasts}
\end{figure}

Figure \ref{fig:forecasts} shows some example forecasts from models over 3 day intervals. The hourly variability is captured by both the LSTM and the linear model as they expect surges and quieter periods over a given day. The LSTM model appears to better describe the different dynamics of weekday versus weekend days. Although both models forecasts on average differ from the observed data around 4 patients, the LSTM model has the lowest prediction error (See Table \ref{tab:results}).

The results, as shown in Table \ref{tab:results}, provide insights into the performance of each model based on metrics such as Mean Squared Error (MSE), Mean Absolute Error (MAE), training time, and prediction time. Among the models, the LSTM3 model exhibited the lowest MSE and MAE, indicating relatively accurate predictions with minimal errors. However, it required a longer training time compared to other models. On the other hand, the Time-varying linear model showed competitive performance with lower MSE and MAE values and relatively shorter training time compared to TBATS and the LSTM configurations. The Naive Approach, which assumes the next observation to be the same as the previous one, has the shortest prediction time since it simply uses the previous week's arrival counts, but had higher MSE and MAE values compared to the other models.

\begin{table}[htbp]
  \centering
  \caption{Forecasting results on the hourly arrival counts of patients at the Rambam Medical Center.}
    \begin{tabular}{lccccc}
      \hline
        \multicolumn{1}{c}{} & \multicolumn{2}{c}{Metrics}  \\
         \hline
          Model              & MSE        & MAE        & Training Time [s]       & Prediction Time [s]\\
        \hline
          Time-varying linear model      & 32.56        & 4.27         & 296.14       & 0.09\\
          \hline
          TBATS               & 34.05         & 4.47 & 5173.41 & 0.03          \\
          \hline
          LSTM3              & 30.43 & 4.17   & 8231.00   &0.39     \\
          \hline
          LSTM3-W              & 32.33 &  4.28  & 6495.20 &0.36        \\
          \hline
          LSTM7              & 37.99 &  4.62   & 5037.49   &0.37     \\
          \hline
          LSTM7-W              & 40.37 & 4.73  & 5069.72 & 0.37         \\
          \hline
           Naive Approach              & 43.81         & 4.94         & NA         & $6.72\times10^{-5}$          \\
          \hline
          RVAR              & 44.56       & 4.96      & NA     & 1.28\\
          \hline
    \end{tabular}
      
  \label{tab:results}
\end{table}

Table \ref{tab:models} provides a summary of the models considered in terms of their features, input horizon, forecast horizon, and explainability. The features used in the models include hourly arrivals and, in some cases, additional variables such as maximum temperature. The input horizon represents the historical data considered for forecasting, while the forecast horizon indicates the duration of the predictions. Notably, some models, such as TBATS and certain LSTM configurations, do not incorporate explainable features, indicating that their predictions are based solely on historical data without explicit interpretability. Overall, the tables offer a comprehensive overview of the forecasting models considered, their characteristics, and their performance metrics, providing insights for decision-makers in healthcare management and resource allocation at the Rambam Medical Center.

\begin{table}[htbp]
  \centering
  \caption{A Summary of the Models Considered}
    \begin{tabular}{lccccc}
      
        \multicolumn{1}{c}{} & \multicolumn{4}{c}{}  \\
         \hline 
          Model              & Features & Input horizon & Forecast Horizon & Explainable\\
        \hline \hline
          Time-varying linear model      & Hourly Arrivals     &7 days    &7 days       &Yes\\
          \hline
          TBATS               &   Hourly Arrivals & Whole history & 1 year & No    \\
          \hline
          LSTM3               &   Hourly Arrivals & 7 days & 3 days  & No    \\
          \hline
          LST3-W              &  \makecell{Hourly Arrivals, \\ Maximum Temperature} & 7 days & 3 days  & No       \\
          \hline
          LSTM7             &   Hourly Arrivals & 7 days & 7 days  & No         \\
          \hline
          LSTM7-W             &  \makecell{Hourly Arrivals, \\ Maximum Temperature} & 7 days & 7 days  & No          \\
          \hline
           Naive Approach        &Hourly Arrivals      &7 days    &7 days       &Yes\\
          \hline
          RVAR       &Hourly Arrivals       &14 days    &7 days    &Yes \\
          \hline
    \end{tabular}
      
  \label{tab:models}
\end{table}


\section{Conclusion} \label{sec: conclusion}
This short forecasting experiment on patient arrivals at an Israeli Medical Center has demonstrated the forecasting capability of two proposed models. The first, a time-varying linear model, makes linear predictions on the number of patients arriving at each hour of a week based on the last 7-day hourly number of arrivals and a set of features that include the day of the week. The second model, a neural network known as an LSTM, learns a non-linear relationship between the previous week of data and a three-day forecasting horizon. Both models demonstrate an ability to capture hourly variability, however, the LSTM model better describes weekly seasonal effects and therefore achieves the lowest prediction error of the two.

Potential future extension of this research is to make the model \ref{eq:ssm} fully online without a training process by explicitly computing the gradients of the likelihood function with respect to all unknown parameters using Kalman filter update, and then by using a recursive online algorithm to optimise the likelihood, for example, \citet{schwartz2019recursive}, to estimate the parameters and make predictions in the same time at each time point. Another future research direction is to explore how to combine the time-varying linear model based method and the LSTM algorithm with different weights to obtain better forecasts, for example, by using the aggregation algorithms \citep{kalnishkan2022prediction} to add different weights to predictions made by different algorithms.



\bibliography{bibliography} 




%
%
%

\newpage

\begin{APPENDICES}
\section{Proof of Lemmas and Propositions}
\subsection{Proof of Theorem \ref{marginal}}
    We use Moment generating function to prove the theorem. We have
    \begin{equation}
    \begin{aligned}
        M_X(t) &= \mathbb{E}[\operatorname{exp}(tX)]=\mathbb{E}[\mathbb{E}[\operatorname{exp}(tX)]|m]\\
        & = \mathbb{E}[\operatorname{exp}(mt+\sigma^2t^2/2)]\\
        &=(\sigma^2t^2/2)\mathbb{E}[\operatorname{exp}(mt)]\\
        & = \sigma^2t^2/2\operatorname{exp}(\theta t+s^2t^2/2)\\
        &=\operatorname{exp}(\theta t+(\sigma^2+s^2)t^2/2),
    \end{aligned}
    \end{equation} which is the moment generating function of a $\mathcal{N}(\theta,\sigma^2+s^2)$.
\subsection{Proof of Proposition \ref{prop1}}
We have that $p(Y_{1},\dots,Y_{T}) = p(Y_{1})\prod_{i=2}^T p(Y_{i}|Y_{i-1})$, where, $$p(Y_{i}|H_i,\beta(t_i))\sim\mathcal{N}(H_i\beta(t_i),\sigma^2),$$ and by the Kalman filter update, we have $$p(H_i\beta(t_i)) \sim\mathcal{N}( H_iB\mu_{i-1|i-1},H_iP_{i|i-1}H_i^\top),$$ and
$$Y_1\sim\mathcal{N}(H_1B\mu_0,\sigma^2+ H_1(BV_0B^\top+Q)H_1^\top).$$Then by Theorem \ref{marginal}, the result follows.

\end{APPENDICES}

\end{document}